\documentclass[pdflatex,sn-mathphys-num]{sn-jnl}

\usepackage{graphicx}%
\usepackage{multirow}%
\usepackage{amsmath,amssymb,amsfonts}%
\usepackage{amsthm}%
\usepackage{mathrsfs}%
\usepackage[title]{appendix}%
\usepackage{xcolor}%
\usepackage{textcomp}%
\usepackage{manyfoot}%
\usepackage{booktabs}%
\usepackage{algorithm}%
\usepackage{algorithmicx}%
\usepackage{algpseudocode}%
\usepackage{listings}%
\usepackage{float}

\theoremstyle{thmstyleone}%
\theoremstyle{thmstyletwo}%

\theoremstyle{thmstylethree}%

\begin{document}

\title[Supervised Coupled Matrix-Tensor Factorization (SCMTF) for Computational Phenotyping of Patient Reported Outcomes in Ulcerative Colitis]{Supervised Coupled Matrix-Tensor Factorization (SCMTF) for Computational Phenotyping of Patient Reported Outcomes in Ulcerative Colitis}


\author*[1]{\fnm{Cristian} \sur{Minoccheri}}\email{minoc@umich.edu}

\author[1]{\fnm{Sophia} \sur{Tesic}}\email{tesics@umich.edu}

\author[1,3,4]{\fnm{Kayvan} \sur{Najarian}}\email{kayvan@med.umich.edu}

\author[1,2]{\fnm{Ryan} \sur{Stidham}}\email{ryanstid@med.umich.edu}

\affil[1]{\orgdiv{Gilbert S. Omenn Department of Computational Medicine and Bioinformatics},  \orgname{University of Michigan}, \orgaddress{\street{1109 Geddes Avenue}, \city{Ann Arbor}, \postcode{48104}, \state{MI}, \country{USA}}}

\affil[2]{\orgdiv{Department of Gastroenterology}, \orgname{University of Michigan}, \orgaddress{\street{1109 Geddes Avenue}, \city{Ann Arbor}, \postcode{48104}, \state{MI}, \country{USA}}}

\affil[3]{\orgdiv{Department of Emergency Medicine}, \orgname{University of Michigan}, \orgaddress{\street{1109 Geddes Avenue}, \city{Ann Arbor}, \postcode{48104}, \state{MI}, \country{USA}}}

\affil[4]{\orgdiv{Department of Electrical Engineering and Computer Science}, \orgname{University of Michigan}, \orgaddress{\street{1109 Geddes Avenue}, \city{Ann Arbor}, \postcode{48104}, \state{MI}, \country{USA}}}


\abstract{\textbf{Background:}
Phenotyping is the process of distinguishing groups of patients to identify different types of disease progression. A recent trend employs low-rank matrix and tensor factorization methods for their capability of dealing with multi-modal, heterogeneous, and missing data. Symptom quantification is crucial for understanding patient experiences in inflammatory bowel disease, especially in conditions such as ulcerative colitis (UC). However, patient-reported symptoms are typically noisy, subjective, and significantly more sparse than other data types. For this reason, they are usually not included in phenotyping and other machine learning methods.

\textbf{Methods:} This paper explores the application of computational phenotyping to leverage Patient-Reported Outcomes (PROs) using a novel supervised coupled matrix-tensor factorization (SCMTF) method, which integrates temporal PROs and temporal labs with static features to predict medication persistence in ulcerative colitis. This is the first tensor-based method that is both supervised and coupled, it is the first application to the UC domain, and the first application to PROs. We use a deep learning framework that makes the model flexible and easy to train.

\textbf{Results:}  The proposed method allows us to handle the large amount of missing data in the PROs. The best model predicts changes in medication 8 and 20 months in the future with AUCs of 0.853 and 0.803 on the test set respectively. We derive interpretable phenotypes consisting of static features and temporal features (including their temporal patterns).

\textbf{Conclusions:} We show that low-rank matrix and tensor based phenotyping can be successfully applied to the UC domain and to highly missing PRO data. We identify phenotypes useful to predict medication persistence - these phenotypes include several symptom variables, showing that PROs contain relevant infromation that is usually discarded.}




\maketitle

\section{Background}\label{sec1}

Phenotyping is the process of identifying clinically meaningful subgroups of patients based on their medical data \cite{loftus2022}. Such groupings are highly valuable for clinicians who aim to understand disease progression and develop personalized treatment plans. Electronic health record (EHR) data, which include laboratory test values, demographics, diagnoses, current prescriptions, and previous medication history, can be complex and tedious for clinicians to manually label and group \cite{Holmes2021}. With an increasing amount and complexity of clinical data available, computational phenotyping, or the process of identifying phenotypes using computer-optimized algorithms, has become a growing field of study \cite{Richesson2016}.

Among these techniques, low-rank matrix and tensor factorization methods stand out for their capability of dealing with multi-modal, heterogeneous, and missing  data \cite{Becker2023}. Generalizing matrix approximation as a sum of rank one terms, the CANDECOMP/PARAFAC (CP) decomposition \cite{Carroll1970, Harshman1970, Hitchcock1927} models a tensor (here, a 3D array) as the sum of rank-one tensors, with each component representing a latent phenotype. For instance, in a $patients$ × $diagnoses$ × $medications$ tensor, CP reveals triplets of factors corresponding to patient groups, co-occurring diagnoses, and prescribed medications that define a phenotype. A key advantage of CP is its essential uniqueness, up to permutation and scaling, which lends stability and reliability to the interpretation of extracted phenotypes, even without additional constraints like orthogonality or independence \cite{Kruskal1977, Sidiropoulos2000}. Constraints like nonnegativity (i.e., requiring the factors to be nonnegative) are usually imposed for interpretability. The property of essential uniqueness is shared by the PARAFAC2 model, which builds on the CP model by relaxing the assumption that all slices must have the same length or alignment across entities along one mode, typically time \cite{han2023}. Thus, PARAFAC2 has mainly been applied in situations with irregular time-series data, such as EHR for clinic visits, where individuals may have varying numbers of visits over time \cite{spartan}. The decision between CP or PARAFAC2 for computational phenotyping depends on the data being used.

A noteworthy advantage of matrix and tensor methods is the possibility of factorizing them jointly using Coupled Matrix-Tensor Factorization (CMTF), effectively merging diverse data such as temporal and static. For example, Afshar et al. proposed the TASTE model, a non-negative PARAFAC2 model that combines a temporal and a static tensor, thus combining both forms of information in phenotypes \cite{Afshar2020}. TASTE is an unsupervised CMTF model that uses a Non-Negativity constrained Least Squares (NNLS) algorithm for optimization. 

Unsupervised matrix and tensor factorization methods have been widely studied as a way of extracting meaningful phenotypes from unstructured, unlabeled datasets \cite{Becker2023}. These phenotypes can be effectively used for downstream tasks such as classification \cite{hodgman}. However, supervised factorization has the potential of yielding better prediction results by linking phenotypes to specific outcomes of interest. The most recent work in this direction -- to our knowledge -- is MULTIPAR \cite{Ren2023}, an extension of PARAFAC2 that is capable of multi-task learning. The MULTIPAR model is a supervised, irregular tensor factorization method that performs gradient-based alternating least squares (ALS) optimization of the decomposition factors. However, MULTIPAR does not incorporate static information into the model. 

Separate from ALS-based approaches, Acar et al. proposed an all-at-once optimization method for coupled matrix and tensor factorizations, calling the method CMTF-OPT \cite{Acar2011}. All-at-once optimization offers key advantages over traditional alternating optimization schemes when solving coupled matrix and tensor factorization problems. Unlike alternating approaches, which update one factor matrix at a time and can suffer from convergence issues or sensitivity to overfactoring, all-at-once optimization like in CMTF-OPT optimizes all parameters simultaneously, leading to more stable and accurate recovery of latent factors. Acar et al. suggest that CMTF with all-at-once optimization can compute factors with low recovery error for data with up to $90\%$ missingness. This key feature is particularly important in situations where data is highly missing, as is often the case with data on symptoms, or Patient-Reported Outcomes (PROs). Additionally, all-at-once optimization naturally integrates with gradient-based learning frameworks, making it well-suited for modern supervised learning settings involving joint training with predictive tasks. Since the proposal of all-at-once optimization for CMTF, no further work has been done to develop the method specifically for computational phenotyping. This is partly due to scalability issues, since computational phenotyping is usually performed on large EHR datasets, making speed a focus of research \cite{Choi2019}. As a consequence, the capabilities of all-at-once optimization remain unexplored. Finally, it is an open problem for tensor methods in computational phenotyping to determine how to deal with different data types (such as lab values and PROs) \cite{Becker2023}. 

\

In this work, we introduce the first Supervised Coupled Matrix-Tensor Factorization (SCMTF) for computational phenotyping (to our knowledge); we additionally show how an all-at-once optimization can be extended to the supervised setting and provides superior performance to ALS methods. Inspired by recommender systems' literature, we introduce bias terms to account for patients' subjectivity and to address the problem of dealing with different data types. We apply our model to a Michigan Medicine dataset consisting of temporal and static features for patients with Ulcerative Colitis. 

Ulcerative colitis (UC) is one of the two major forms of inflammatory bowel disease (IBD) and primarily affects the mucosal layer of the colon \cite{Massano2025, Matsuoka2018, Kobayashi2020}.
The disease is estimated to affect about 1.25 million Americans \cite{Lewis}. Treatment for UC is designed to achieve remission before transitioning to long-term maintenance therapy. Remission in UC is defined as the period when a patient’s symptoms resolve and the disease is not interfering with their daily life \cite{Moss2014}. As such, PROs are highly relevant in treatment decisions.
Successful UC treatment is characterized by long-term maintenance of an unchanged medication regimen. Therefore, there is significant clinical interest in identifying characteristics shared by patients who experience medical treatment persistence following the initiation of a novel UC medication regimen. Though PROs are of direct relevance to UC and many other diseases, they are seldom used in computational phenotyping, as the data are typically noisy, more subjective, and significantly more sparse when compared to traditional EHR data.

Given the reduced size and the structured nature of our dataset compared to traditional datasets used in tensor-based phenotyping, we propose a Supervised CMTF model for that integrates static and temporal EHR and PRO data using nonnegative CP decomposition with bias terms for temporal features and patients and all-at-once optimization, in order to predict the long-term success of treatment.

\

 The novelty of this work is therefore as follows:

 1) we introduce and validate the first Supervised Coupled Matrix-Tensor Factorization (SCMTF) framework for computational phenotyping;

 2) we extend all-at-once optimization to the supervised setting to seamlessly learn a constrained coupled decomposition and train a neural network classifier;

 3) we apply phenotyping to a new type of EHR data that includes noisy and highly missing PROs, and introduce bias terms to account for patients' subjectivity and to address the problem of dealing with different data types (labs and PROs);

 4) we apply tensor-based computational phenotyping to a domain that has not been studied so far: inflammatory bowel disease.

\section{Methods}\label{sec2}

\subsection{Dataset}

A retrospective single center dataset was developed containing 2,303 adult UC patients using EHR data from between 2018 and 2023.  UC diagnosis was confirmed using a validated method requiring 3 separate outpatient encounters where UC was the primary diagnostic code, plus at least one IBD related medication.  Selection criteria included the initiation of a new biologic or advanced therapy as confirmed by medication dispensation or infusion records. In addition, all subjects were required to have moderate to severe UC based on documented endoscopic activity within the 6 months prior to medication start. Included subjects were required to have at least 18 months of follow up prior to and at least 36 months of follow up after new medication start (or a medication failure event), as determined by gastroenterologist specialty visit.  Exclusion criteria included subjects not completing induction therapy phase, with an ostomy, ileoanal pouch anastomosis, those using enteral or parenteral supplemental nutrition, and those using chronic antibiotic therapy.

EHR data includes static demographic variables of age at the time of new medication start, sex, disease duration, disease location (pancolitis, left colon, proctitis), and endoscopic severity prior to new medication start.  Medication use history included prior IBD medication exposures of more than two months based on medication dispensation or infusion records, both by medication class and specific medication type.  The minimum 2 months medication use criteria (e.g. completion of induction) aids in avoiding patients discontinuing a new therapy due to intolerance or allergy. Laboratory data collected from EHR records included inflammatory biomarkers of C-reactive protein and fecal calprotectin. Complete blood counts, specifically hemoglobin, hematocrit, platelet, and white blood cell counts were included.  Blood chemistries included blood urea nitrogen and albumin.
Patients electronically submit the UC-PRO/SS as part of routine outpatient UC care at our institution. The UC-PRO/SS is a validated instrument used in clinical trials and care that has shown high reproducibility (ICC=0.81) and very good correlation with the Partial Mayo Score (r=0.79).  The UC-PRO/SS is comprised of 6 modules including bowel signs and symptoms, abdominal symptoms, systemic symptoms, coping strategies, daily life impact, and emotional impact.  All scores are patient reported using a 0-5 Likert-like severity scale. The primary outcome was medication persistence, as determined by confirmation of ongoing medication use vs. observed medication discontinuation or a surgical colectomy.

\begin{figure}[h]
\caption{Description of dataset organization}
\centering
\includegraphics[scale=0.2]{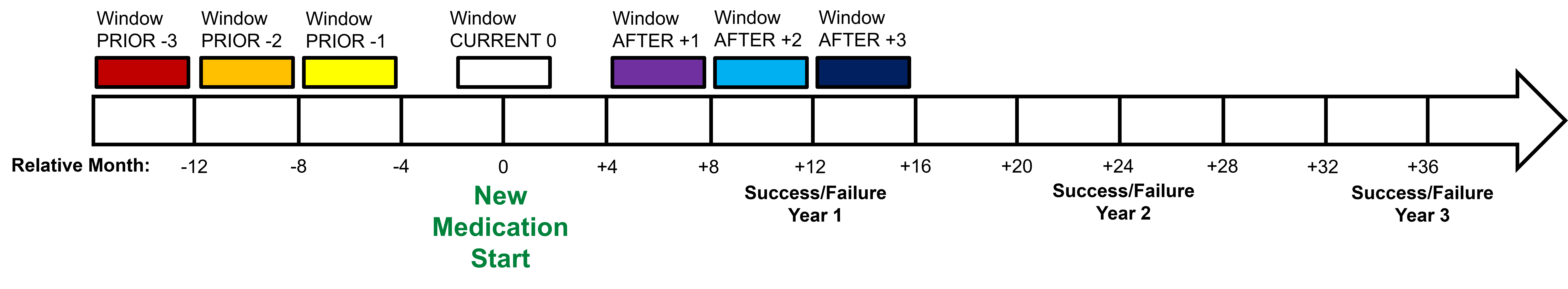}
\label{data}
\end{figure}

Temporally sensitive data (e.g. PRO symptoms and laboratory data) was organized in 4 month window blocks, anchored to the new medication start date as time “0” (figure \ref{data}). Timepoint windows included -16-12, -12-8, -8-4 month blocks prior to new medication start and +4-8, +8-12, and +12-16 month windows following medication start; there was a total of 7 timepoint windows considering the baseline medication start window.

Nearly all subjects had some missing data. It was impossible to consider a subset of patients with mostly non-missing data due to small size. For example, there were only 64 patients with less than 30$\%$ missing values. To address the patients missing nearly all data, we removed all patients with $\geq$ 90$\%$ missing values across all features. We imputed only static variables using mean imputation, and fill in missing binary outcomes as failures for medication persistence. We standardized each static feature and standardized each timepoint. To balance lab values, many of which take very high scale values, with PROs, we then normalized each timepoint [0,1]. We also performed normalization to range [0,1] for the static values. We split the data into training, validation, and testing sets using a 60-20-20 split stratified on both outcomes. 

\subsection{Coupled Matrix-Tensor Factorizations}

Coupled Matrix-Tensor Factorization (CMTF) is a mathematical framework used to jointly analyze and decompose heterogeneous datasets, such as matrices and tensors, that share common dimensions or modes. It integrates matrix factorization and tensor factorization techniques to uncover shared latent structures across datasets. The factorization is typically formulated as an optimization problem where the goal is to minimize reconstruction errors for both the tensor and matrices using techniques like alternating least squares or gradient-based methods. We introduce here the basic notation and refer the reader to tensor methods surveys (e.g., \cite{tensor-survey}) for additional details.

A matrix $M$ is a two-dimensional array whose entries can be described by two indices as $M_{ij}$, where $M_{ij}$ is the entry on the $i-$th row and $j-$th column. A $3-$way tensor $T$ is a three-dimensional array whose entries can be described by three indices as $T_{ijk}$.  

Tensor factorizations are a generalization of matrix factorization techniques. We will consider one type of tensor factorization -- the Canonical-Polyadic (CP) decomposition. Given matrices $A$, $B$, and $C$ of shapes $a\times r$, $b\times r$, and $c\times r$ respectively (note the same number of columns), we can construct a tensor $T=[A,B,C]$ whose entry $T_{ijk}$ equals $$\sum_{s=1}^rA_{is}B_{js}C_{ks}.$$ We say that $[A,B,C]$ is a CP factorization of $T$ and that $A,B,C$ are factor matrices of $T$.

In applications, we usually go the other way around: given a 3-way array $T$ of shape $a \times b \times c$ and a rank $r$, we would like to find matrices $A$, $B$, and $C$ of shapes $a\times r$, $b\times r$, and $c\times r$ respectively, such that $[A,B,C]$ is as close as possible to $T$. This is called a CP approximation of $T$ rank $r$, and it generalized the rank $r$ approximation of a matrix $M$ given by $AB^t$ ($B^t$ is the transpose of $B$). Generally, as close as possible means minimizing the Frobenius norm $\|T-[A,B,C]\|^2$, where $\|T\|^2$ is the sum of the squared entries of $T$.

To enhance interpretability, uniqueness, and numerical stability, we normalize the factor matrices column-wise and introduce weights $w_i=\|a_i\|\cdot \|b_i\| \cdot \|c_i\|$ for $i=1,\ldots,r$, collected in a vector $w=(w_1,\ldots,w_r)$ of length $r$. The resulting factorization is denoted $[w;A,B,C]$ and its entry $(i,j,k)$ equals $$\sum_{s=1}^r w_iA_{is}B_{js}C_{ks}.$$

\begin{figure}[h]
\caption{Coupled Matrix-Tensor Approximation}
\centering
\includegraphics[scale=0.3]{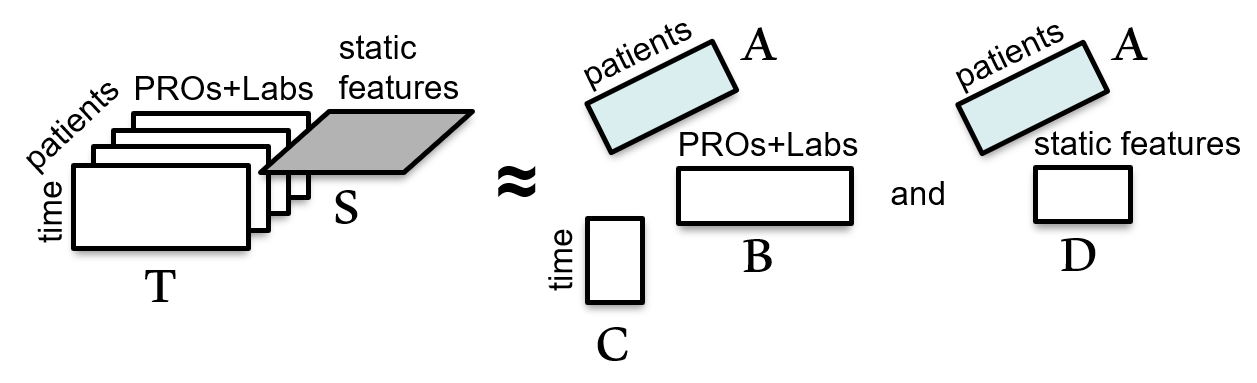}
\label{CTMF}
\end{figure}

We structure our data into the form of a tensor $T$ with modes (patients) $\times$ (labs and PROs) $\times$ (time) of size $a \times b \times c$, and a matrix $M$ with modes (patients) $\times$ (static features). The tensor and the matrix align along the patient mode, so we perform CMTF, where a term $[A,B,C]$ is used to approximate the tensor and the term $AD^t$ is used to approximate the matrix, for factor matrices $A$, $B$, $C$, and $D$ of shapes $a\times r$, $b\times r$, $c\times r$, and $d\times r$. The decomposition is depicted in figure \ref{CTMF}. For notational consistency, we denote $AD^t$ by $[w; A,D]$, whose entry $(i,j)$ equals $$\sum_{s=1}^r w_s A_{is}D_{js}.$$ In the coupled factorization, all matrices $A,B,C,D$ are normalized column-wise, and $w_i=\|a_i\|\cdot \|b_i\|\cdot \|c_i\|\cdot \|d_i\|$, where $w=(w_1,\ldots,w_r)$ is still a vector of length $r$.

The matrix $A$ is the same in the approximations for both the tensor and the matrix, which allows us to combine information from both sets of data at the patient level. The factors $A$, $B$, $C$, and $D$ of this joint decomposition are parametrized matrices learned during training of the model. Of the several choices of algorithms for learning the matrices $A$, $B$, $C$, and $D$, we chose a gradient method because it allows us to easily include a supervised term along with additional constraints, which we will discuss later.

\subsection{Proposed Supervised Coupled Matrix-Tensor Factorization (SCMTF) model}

We developed a neural network-based, supervised, coupled matrix-tensor factorization model. We define a phenotype as formed by the $i-$th columns $a_i$, $b_i$, $c_i$, and $d_i$ of each of the matrices $A$, $B$, $C$, and $D$ for $i=1, \ldots, r$. The entries of $a_i$ quantify for each patient to what extent they belong to the $i-$th phenotype. We therefore refer to $A$ as the patient-phenotype membership matrix. The entries of $d_i$ reveal to what extent static features appear in the $i-$th phenotype. The entries of $b_i$ reveal to what extent temporal features appear in the $i-$th phenotype. The entries of $c_i$ reveal the temporal pattern of the $i-$th phenotype across the data.

In order to make the phenotypes relevant to the outcomes of interest, we included a classifier that is jointly trained with the model. Given our patient-phenotype membership matrix $A$, we train a classifier on this matrix to predict the labels.

During factorization, to handle missing data in the tensor $T$, we can introduce a binary mask $\Omega$ (of the same shape as $T$) whose entry $\Omega_{ijk}$ equals $1$ if $T_{ijk}$ is known and $0$ otherwise. This way, the entry-wise product $\Omega * T$ simply replaces the unknown entries of $T$ with zeros. In approximating $T$ with $[w;A,B,C]$ we will only consider known values, i.e., we seek to minimize $\|\Omega * (T-[w;A,B,C])\|^2$. However, note that $[w;A,B,C]$ is a tensor whose entries are completely known -- for this reason, a CP approximation can be used to impute missing values.

The coupled matrix-tensor decomposition problem (with missing values in the tensor) can be written as $${\arg\min}_{w,A,B,C,D} \|\Omega*(T-[w;A,B,C])\|^2 + \|[w;A,D]\|^2.$$ To enhance interpretability of the phenotypes and uniqueness of the decomposition, it is useful to impose nonnegativity constraints, which means restricting $w,A,B,C,D$ to have nonnegative entries.

Given the subjective nature of PROs, we add bias terms to account for the fact that a patient might generally score symptoms higher than another one (with the same underlying gravity of the disease) and the fact that a symptom might generally be scored higher by patients because it has a stronger effect on the patient's life. In fact, we add bias terms for all temporal features (i.e., lab values as well) to account for the fact that lab values are continuous and have strictly positive baselines whereas PROs are discrete and significantly more sparse. This discrepancy is known to cause problems in factorization models and has led to the use of different losses for different types of data \cite{GLRM} and is an open problem in computational phenotyping \cite{Becker2023}. The implementation results in introducing a tensor $B_{features}$ with constant slices for each temporal feature and a tensor $B_{patients}$ with constant slices for each patient. The values of each slice are learnable parameters of a vector $b$; for each feature slice of $B_{features}$ and each patient slice of $B_{patients}$, the values are tied to stay constant during backpropagation; the bias tensors are added to $[w;A,B,C]$.

We can encourage phenotypes to contain fewer features by adding $\ell_1-$penalty terms ($\ell_1(X)$ for a matrix $X$ is the sum of the absolute values of the entries of $A$) in the form of a sparsity term: $$L_{sparse}(A,B,C,D)= \ell_1(A) + \ell_1(B) + \ell_1(C) + \ell_1(D).$$ Since the $\ell_1-$norms in the sparsity term are not differentiable, we use proximal gradients \cite{murray2019} to update $A,B,C$ and $D$ .

Finally, we include the loss $L_{cl}(A,Z)$ of a neural network classifier with parameters $Z$ applied to the matrix $A$. The classifier is a 2-layer fully connected network with $10$ hidden nodes and $2$ output nodes (one node per outcome); the first layer includes batch normalization and a ReLU activation function.

Therefore, the final optimization problem is: $${\arg\min}_{w,A,B,C,D,b} (1-\lambda)(L_{rec}(w,A,B,C,D,b)) + \lambda L_{cl}(A,Z),$$ where $L_{rec}(w,A,B,C,D,b)$ is the reconstruction loss and it is equal to $$\|\Omega*(T-[w;A,B,C]-B_{features}-B_{patients})\|^2 + \|[w;A,D]\|^2 +L_{sparse}(A,B,C,D),$$ and $\lambda$ is a hyperparameter used to balance reconstruction loss and classifier loss.

Since this is a constrained optimization problem, with the constraints given by conditions ${w,A,B,C,D \geq 0}$, we adopted a projected gradient descent strategy: during the forward pass, we compute the total loss $$(1-\lambda)(L_{rec}(w,A,B,C,D,b)) + \lambda L_{cl}(A,Z)$$ and then backpropagate and clamp the values of $w,A,B,C,D$ to ensure nonnegativity. The implementation uses TensorLy \cite{tensorly} for basic tensor operations and is available on GitHub at https://github.com/Minoch/SCMTF.git

\section{Results}

We initialize the factor matrices $A$, $B$, $C$, and $D$ and the bias vector $b$ randomly in the interval $[0,1]$. We also initialize the weight vector $w$ to all ones. We define two separate optimizers to update the decomposition and classifier parameters. Specifically, we define an Adam optimizer to update $A$, $B$, $C$, $D$, $w$, and $b$, and a Stochastic Gradient Descent (SGD) optimizer to update the parameters of the neural network classifier jointly trained during decomposition. Since the $\ell_1-$norms in the sparsity term are not differentiable, we additionally use proximal gradients to update $A,B,C$ and $D$, as implemented in \cite{murray2019}. To further enhance the training process, we apply a learning rate scheduler to both optimizers. The scheduler decays the learning rate by a factor of $0.8$ every $1000$ steps for both the Adam and SGD optimizers.  We optimized hyperparameters for the model on a grid space reported in table \ref{hyper}. To assess how well we dealt with missing data, we randomly removed a fraction of $0.05$ of known data and let the model impute the removed values. Using missing value imputation and AUC on the validation set as performance measures, we find the best hyperparameters to be rank $r=28$, balance term $\lambda=0.7$, learning rate $lr=0.01$, and penalty $\ell_{1}=0.001$. We obtained a Mean Absolute Error (MAE) of $0.145$ and a Residual Mean Squared Error (RMSE) of $0.197$ in comparing removed real values and imputed values. 

\begin{table}[h]
\centering
\caption{Hyperparameter Grid Search Space}
\begin{tabular}{@{}ll@{}}
\toprule
\textbf{Hyperparameter} & \textbf{Values} \\ \midrule
Rank $r$ & \{10, 12, 14, 16, 18, 20, 22, 24, 26, 28, 30, 32\} \\
Balance term $\lambda$ & \{0.5, 0.6, 0.7, 0.8, 0.9\} \\
Learning rate $lr$ & \{0.001, 0.01, 0.1\} \\
Penalty $\ell_{1}$ & \{0, 0.001, 0.01, 0.1\} \\
\bottomrule
\end{tabular}
\label{hyper}
\end{table}

After training is complete, we calculate several sets of metrics, summarized in table \ref{tab2}.
Since the rank is $28$, we have $28$ raw phenotypes. Phenotype membership of each patient is contained in the patient-phenotype matrix $A$, which we can consider as a feature matrix with $28$ features. First, we evaluate the model on the test set and report classifier metrics using the same neural network (NN) classifier used during training. Second, we train a Random Forest (RF) model on the matrix $A$ and evaluate prediction on the test set. We train a separate model for each outcome, using the same RF parameters for both, and evaluate both classifiers on the test set from the initial model. Finally, we identify the top phenotypes based on feature importance of the RF classifier setting a threshold at $0.10$. We obtain with 3 distinct phenotypes: one of the phenotypes, phenotype 24 is highly important for both outcome labels; phenotype 16 is of high importance for the outcome of medication persistence at year 2, and phenotype 27 is highly important for the outcome at year 3. To determine how informative this smaller set of phentypes is, we trained a second set of RF models, one for each outcome, on these top 3 phenotypes (RF top).

The best model performance on the test set for medication persistence at 2 years has an AUC of $0.853$, and the best model performance for medication persistence at 3 years has an AUC of $0.803$.

\begin{table}[h]
\caption{Classification performance on the test set of the neural network classifier built in the proposed model (``NN"), of RF classifiers trained on top of the proposed model (``RF"), and of RF classifiers trained on the most relevant phenotypes (``RF top")}\label{tab2}
\begin{tabular*}{\textwidth}{@{\extracolsep\fill}lcccccc}
\toprule%
& \multicolumn{3}{@{}c@{}}{Year 2} & \multicolumn{3}{@{}c@{}}{Year 3} \\\cmidrule{2-4}\cmidrule{5-7}%
 & RF & RF top & NN & RF & RF top & NN \\
\midrule
AUC  & \bf{0.853} & 0.812  & 0.807 & \bf{0.803}  & 0.772  & 0.730\\
F1  & \bf{0.832}  & 0.805  & 0.737  & 0.672 & \bf{0.672}  & 0.509\\
Precision  & \bf{0.774} & 0.760 & 0.771  & 0.646  & \bf{0.655} & 0.603\\
Recall  & \bf{0.899} & 0.855  & 0.706  & \bf{0.699} & 0.689 & 0.440\\
\botrule
\end{tabular*}
\label{RF}
\end{table}

Given the novel nature of SCMTF, there are no standard models that can be used for comparison. Given the high percentage of missing data in PROs, classical imputation methods were not feasible and it was also not possible to consider a subset of patients with mostly known PROs. Therefore, running standard machine learning models for comparison was not possible. Additionally, tensor factorization methods for computational phenotyping in the literature are either not coupled or not supervised. The closest model being MULTIPAR, we modified our code to create a version that would be a fair comparison with it. MULTIPAR solves the optimization problem via gradient descent but optimizing each factor matrix alternatingly, in ALS fashion. It also trains a logistic regression classifier at each iteration (instead of jointly with the decomposition) and adds the trained classifier loss to the total loss. We call this version ``ALS" and compare results on the test set with our method, ``all-at-once". For an even more fair comparison, we also add bias to this adapted MULTIPAR model (``ALS with bias"). Results are shown in table \ref{tab3}. In terms of missing value imputation, the ALS model shows worse imputation: MAE $= 0.271$, RMSE $= 0.372$. However, after adding bias the imputation performance is analogous to the all-at-once model.

\begin{table}[h]
\caption{Classification performance on the test set of the proposed model (``all-at-once"), a fair version of the MULTIPAR model using alternating least squares (``ALS"), and a fair version of the MULTIPAR model using alternating least squares with bias terms like in the proposed model (``ALS with bias") }\label{tab3}
\begin{tabular*}{\textwidth}{@{\extracolsep\fill}lcccccc}
\toprule%
& \multicolumn{3}{@{}c@{}}{Year 2} & \multicolumn{3}{@{}c@{}}{Year 3} \\\cmidrule{2-4}\cmidrule{5-7}%
 & ALS & ALS with bias & all-at-once & ALS & ALS with bias & all-at-once \\
\midrule
AUC  & 0.656 & 0.719  & \bf{0.807} & 0.643  & 0.700 & \bf{0.730}\\
F1  & 0.690  & \bf{0.764}   & 0.737  & 0.617  & \bf{0.678}  & 0.509\\
Precision  & 0.676 & 0.713   & 0.\bf{771}  & 0.548  & 0.593 & \bf{0.603}\\
Recall  & 0.706 & \bf{0.823}  & 0.706  & 0.705 & \bf{0.793} & 0.440\\
\botrule
\end{tabular*}
\label{ALS}
\end{table}

\section{Discussion}

The imputation results suggest that the model performs well in recovering missing data, even under a controlled test where $5\%$ of known values were randomly removed. With an MAE of $0.145$ and an RMSE of $0.197$, the model demonstrates good overall accuracy within the normalized $[0,1]$ space. For PROs with five discrete, ordinal levels, an MAE of $0.145$ corresponds to being, on average, less than one level away from the true value, which is an encouraging result given the inherent subjectivity and variability in self-reported symptom data. The total set of data entering the model had approximately $75.7\%$ missing values, with the PROs subset of the data being approximately $88.3\%$ incomplete. We believe that the very high percentage of missing data is a limitation of our work. Particularly for the PROs, the inherent noise and volatility of the subjective survey results may be a reason for the model not achieving stronger performance.

The SCMTF method enables us to implement a final classifier outside the training loop. We chose RF for its reliability as a non-linear model with generally good performance. Our results from the RF models for the separate outcomes add more detail to our interpretation about the quality of the phenotype set to capture latent patient groupings. Specifically, the set of $28$ phenotypes achieves good predictive performance for medication persistence at year 2, with an AUC of 0.853, high recall (0.899), and a solid F1 score (0.832), indicating effective identification of relevant cases. We also demonstrate good overall predictive performance for medication persistence at year 3, with an AUC of 0.803. While the NN classifier within the model has significant AUCs, the performance of RF on the extracted phenotypes shows that the phenotypes captured relevant information. The performance of RF on three phenotypes (RF top) being a close second suggests in particular that few phenotypes contain most information.

The comparison with an ALS version of the model shows significant improvements in prediction (0.807 versus 0.656 for year 2, 0.730 versus 0.643 for year 3) as well as missing value imputation. The gap becomes more narrow once bias is introduced in the ALS model, and the missing value imputation error becomes the same, suggesting how relevant bias terms are, in particular to estimate missing values. While recall is higher for the ALS with bias model compared to the all-at-once model, the AUC scores indicate a better overall discriminative power. As previously discussed, there is a lack of standard algorithms for comparison with the proposed methods, but we believe this analysis suggests an improvement of the closest existing models and a benefit of gradient-based, all-at-once optimization.

\begin{table}[h]
\centering
\caption{Fraction of positive class by phenotype}
\begin{tabular}{@{}llr@{}}
\toprule
\textbf{Phenotype} & \textbf{Year 2} &\textbf{Year 3}\\ \midrule
16 & 0.823 & 0.615 \\
24 & 0.851 & 0.701 \\
27 & 0.837 & 0.743 \\
\bottomrule
\end{tabular}
\label{fraction}
\end{table}

\begin{figure}[h]
\caption{Temporal pattern of the top phenotypes}
\label{temporal}
\centering
\includegraphics[scale=0.4]{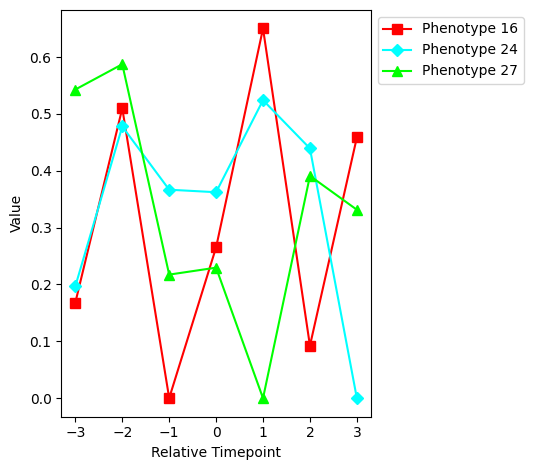}

\end{figure}

One advantage of tensor-based methods is their interpretability. We show in table \ref{top3} the top three phenotypes identified by the RF model: 16, 24, and 27. Phenotype $k$ is obtained from the $k-$th column of matrices $B$, $C$, and $D$; for each temporal and static feature, the column entry of matrices $B$ (temporal) and $D$ (static) shows the membership level of that feature to the phenotype; the $k-$th column of matrix $C$ shows the temporal pattern of the phenotype over time. Phenotypes can be thought of as (possibly overlapping) clusters of patients, grouped together based on the value of certain features, the pattern of these features over time, and their outcome. For ease of interpretation, as customary, we only show the most relevant features (contribution $>0.2$) for each phenotype. We can visualize the temporal patterns of the top three phenotypes using the factor matrix $C$ as shown in figure \ref{temporal}. The temporal factor tells us how a phenotype's expression for the temporal features changes over time - for example, whether a phenotype's temporal features trend is chronic (persists over time) or episodic (appears at intervals). Different phenotypes may have distinct temporal signatures and by comparing them we can determine their co-occurrence or lack thereof.

Phenotype 24 is identified by the RF model as relevant for both outcomes; additionally, phenotype 16 is identified as relevant for year 2 outcome, and phenotype 27 is identified as relevant for year 3 outcome. We see that among the static features, a change of medication at year 1, weight and age are relevant features across phenotypes, as we might expect. However, current medications and one specific prior medication (among all possible ones), Prior$\_$med$\_$ADA, are also relevant in determining long term medication persistence. With respect to temporal features, we see that the model identifies symptoms are relevant for treatment persistence. In particular, arthritis-like symptoms are the most relevant and appear across phenotypes, whereas DA$\_$ symptoms related to activities only appear in phenotype 16 and therefore are marginally relevant and only in the short term (year 2 outcome).  However, these temporal features have different patterns: phenotype 24 is generally consistent across timepoints, indicating patients with consistent symptoms of pain, bloathing and arthritis, as well as high albumin lab values; phenotypes 16 and 27 are more episodic. In particular, phenotype 27 identifies a pattern where temporal features are significant early on, the symptoms improve and the lab values decrease, but eventually symptoms worsen and lab values increase for the last two time points.

\begin{figure}[h]
\caption{Lab$\_$Hgb average values for member patients}
\centering
\includegraphics[scale=0.4]{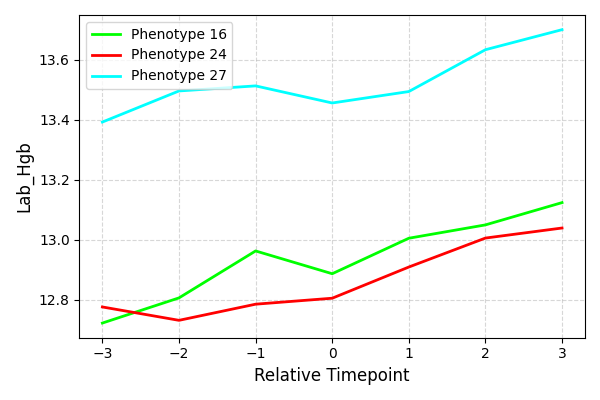}
\label{hgb}
\end{figure}

 We can see in table \ref{fraction} which fraction of patients that belong to a given phenotype have a positive class for each of the two outcomes. While the best predictive results are obtained treating phenotype membership as features and running a RF model on it, the table shows how the phenotypes tend to capture patients with a specific outcome, especially for Year 2 change of medication, and they might therefore have clinical relevance. Additionally, we can identify feature trends of a phenotype by plotting the average feature values for patients that belong to that phenotype. Given the high percentage of missing PROs, these plots cannot be considered reliable for symptoms; however, we can see in figure \ref{hgb} how patients in phenotype 27 have a constantly higher Lab$\_$Hgb values

\begin{table}[h]
\label{top3}
\centering
\caption{Top three phenotypes identified by RF among the 28 learned by the SCMTF model}
\begin{tabular}{@{}llr@{}}
\toprule
\multicolumn{3}{c}{\textbf{Phenotype 16}} \\
\midrule
\textbf{Feature Type} & \textbf{Feature} & \textbf{Contribution Score} \\
\midrule
Temporal & symptom\_arthritis & 0.356 \\
Temporal & feeling\_angry & 0.339 \\
Temporal & DA\_enjoyment & 0.337 \\
Temporal & DA\_leavinghome & 0.310 \\
Temporal & feeling\_worried & 0.310 \\
Temporal & Lab\_CRP & 0.282 \\
Temporal & Lab\_Wbc	& 0.258\\
Temporal & symptom\_weak & 0.254 \\
Temporal & DA\_planning & 0.236 \\
Temporal & feeling\_alone & 0.222 \\
Temporal & DA\_travel & 0.202 \\
Static & New\_Med\_Outcome\_yr1 & 0.726 \\
Static & WeightKG2 & 0.366 \\
Static & age & 0.342 \\
Static & Current\_med\_MESAL & 0.286 \\
Static & Current\_med\_THIO	& 0.272 \\
Static & Prior\_med\_ADA & 0.216 \\
\midrule
\multicolumn{3}{c}{\textbf{Phenotype 24}} \\
\midrule
Temporal & Lab\_Alb & 0.455 \\
Temporal & symptom\_arthritis & 0.396 \\
Temporal & symptom\_bloating & 0.344 \\
Temporal & symptom\_pain & 0.336 \\
Temporal & Lab\_PLT & 0.277 \\
Temporal & Lab\_CRP & 0.273 \\
Temporal & feeling\_frustrated & 0.238 \\
Static & New\_Med\_Outcome\_yr1 & 0.779 \\
Static & WeightKG2 & 0.363 \\
Static & age & 0.328 \\
Static & Prior\_med\_ADA & 0.290 \\
\midrule
\multicolumn{3}{c}{\textbf{Phenotype 27}} \\
\midrule
Temporal & Lab\_Hgb & 0.593 \\
Temporal & symptom\_arthritis & 0.549 \\
Temporal & symptom\_tired & 0.317 \\
Temporal & Lab\_Alb & 0.237 \\
Temporal & feeling\_alone & 0.234 \\
Static & New\_Med\_Outcome\_yr1 & 0.686 \\
Static & WeightKG2 & 0.380 \\
Static & age & 0.352 \\
Static & Current\_med\_THIO & 0.349 \\
Static & Current\_med\_MESAL & 0.323 \\
\bottomrule
\end{tabular}
\end{table}

\begin{figure}[H]
\caption{Distribution of patients' bias}
\label{patient_bias}
\centering
\includegraphics[scale=0.5]{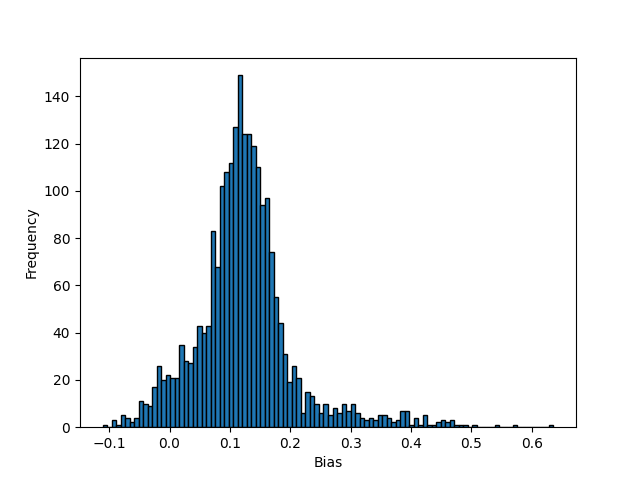}
\end{figure}

\begin{figure}[H]
\caption{Bias values of temporal features}
\label{feature_bias}
\centering
\includegraphics[scale=0.4]{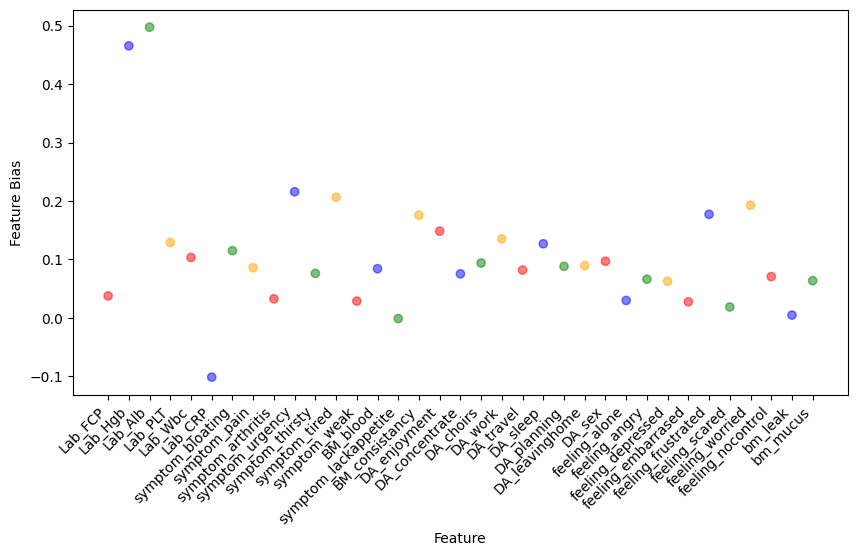}
\end{figure}

 Finally, we can observe the values of learned patient biases (figure \ref{patient_bias}) and feature biases (figure \ref{feature_bias}). Biases for patients follow a gaussian distribution as we would expect, with a positive mean indicating that some patients have generally higher/lower values, especially in PROs reporting. Biases for features identify two lab values as prevalent - Lab$\_$Hgb and Lab$\_$Alb. Using bias allows the phenotypes to account for the systematic baseline of the lab values while still including them as relevant.


\section{Conclusions}
The goal of this work was to develop a supervised CMTF method using all-at-once optimization for computational phenotyping of UC patients. This novel SCMTF method demonstrates good performance in predicting medication persistence at 2 and 3 years after the initiation of a new medication regimen.
In particular, the model is able to combine static and temporal features and to deal with noisy and highly missing PROs data. Additionally, the method produces interpretable phenotypes that can be used to understand the model predictions and to identify groups of patients with similar patterns. Interestingly, the phenotypes include several symptoms, showing that the proposed method can extract meaningful information from PROs.

However, there are several limitations to our work. The PROs are inherently noisy and highly missing, and although we demonstrated our method's ability to extract valuable information from the data to predict medication persistence, less missing data would allow us to perform a more robust evaluation and to analyze the phenotypes more thoroughly from the point of view of PROs (for example, analyzing their temporal trends for each phenotype). To draw broader conclusions, particularly for clinical application of these phenotypes, we would need a more comprehensive set of data from different institutions. We used a fairly representative patient age pool in this study, but may benefit from additional patients at the youngest and oldest ends of the distribution to maximize the model's ability to generalize to future patients.

\backmatter

\section*{List of abbreviations}

ALS - Alternating Least Squares

\noindent AUC - Area Under the Receiver operating characteristic curve

\noindent CMTF - Coupled Matrix-Tensor Factorization

\noindent CP - CANDECOMP/PARAFAC 

\noindent EHR - Electronic Health records

\noindent IBD - Inflammatory Bowel Disease

\noindent MAE - Mean Absolute Error

\noindent NN - Neural Network

\noindent PROs - Patient-Reported Outcomes

\noindent RF - Random Forest

\noindent RMSE - Residual Mean Squared Error

\noindent SCMTF - Supervised Coupled Matrix-Tensor Factorization

\noindent SGD - Stochastic Gradient descent

\noindent UC - Ulcerative Colitis

\section*{Declarations}

\begin{itemize}
\item Funding: Research reported in this publication was supported by the Center for Data-Driven Drug Development and Treatment Assessment (DATA), an industry-university cooperative research center partially supported by the U.S. National Science Foundation under the award number 2209546, and by DATA industry partners.
\item Conflict of interest/Competing interests: Not applicable.
\item Consent for publication: Not applicable.
\item Availability of data and materials: The datasets generated and/or analyzed during the current study are not publicly available.
\item Authors' contributions: data acquisition - RS; funding acquisition - CM, RS, KN; project ideation - CM, RS; data preprocessing: RS, CM, ST; software: CM, ST; first draft: CM, ST, RS; final revision: CM, RS, KN.
\end{itemize}

\begin{appendices}

\section{Features description}\label{secA}

We add a complete list of static and temporal features used as well as their description in table \ref{appendix}.

\begin{table}[h]
\centering
\caption{Dataset Feature Descriptions}
\label{appendix}
\begin{scriptsize}
\begin{tabular}{@{}ll@{}}
\toprule
\textbf{Feature} & \textbf{Description} \\ \midrule
$Prior\_med\_IFX$ & indicates prior use of this biologic medication before new med start date\\
$Prior\_med\_ADA$ & indicates prior use of this biologic medication before new med start date\\
$Prior\_med\_CZP$ & indicates prior use of this biologic medication before new med start date\\
$Prior\_med\_GOL$ & indicates prior use of this biologic medication before new med start date\\
$Prior\_med\_TOF$ & indicates prior use of this biologic medication before new med start date\\
$Prior\_med\_UPA$ & indicates prior use of this biologic medication before new med start date\\
$Prior\_med\_VDZ$ & indicates prior use of this biologic medication before new med start date\\
$Prior\_med\_UST$ & indicates prior use of this biologic medication before new med start date\\
$Current\_med\_THIO$ & indicates current usage of these medications at time of new medication start\\
$Current\_med\_MESAL$ & indicates current usage of these medications at time of new medication start\\
$BMI2$ & Demographic \\
$WeightKG2$ & Demographic \\
$age$ & Demographic \\
$New\_Med\_Outcome\_yr1$ & Medical treatment persistance for 1 years\\
$Lab\_FCP$ & Fecal calprotectin value,
 measure of gut inflammation\\
$Lab\_Hgb$ & Hemoglobin value, measure of chronic inflammation\\
$Lab\_Alb$ & Albumin value, measure of chronic inflammation\\
$Lab\_PLT$ & Platelet value, measure of chronic inflammation\\
$Lab\_Wbc$ & White blood cell count value, measure of general inflammation\\
$Lab\_CRP$ & C reactive protein value, measure of general inflammation\\
$symptom\_bloating$ & In the past 24 hours, did you feel bloating in your belly? If so, how severe?\\
$symptom\_pain$ & In the past 24 hours, did you feel pain in your belly? If so, how severe? \\
$symptom\_arthritis$ & In the past 24 hours, did you feel pain in your knees, hips, and/or elbows?\\
&If so, how severe?\\
$symptom\_urgency$ & In the past 24 hours, did you feel the need to have a bowel movement\\
& right away? If so, how severe?\\
$symptom\_thirsty$ & In the past 24 hours, did you feel thirsty?  If so, how severe?\\
$symptom\_tired$ & In the past 24 hours, did you feel tired?  If so, how severe?\\
$symptom\_weak$ & In the past 24 hours, did you feel weak?  If so, how severe?\\
$BM\_blood$ & In the past 24 hours, did you have blood in your bowel movements? \\
&If so, how often? \\
$symptom\_lackappetite$ & In the past 24 hours, did you lack an appetite? If so, how severe? \\
$BM\_consistancy$ & In the past 24 hours, did you leave home for any reason?\\
$DA\_enjoyment$ & In the past seven days, how much did your disease\\
& interfere with activities you do for enjoyment?\\
$DA\_concentrate$ & In the past seven days, how much did your disease
\\ &interfere with your ability to concentrate?\\
$DA\_choirs$ & In the past seven days, how much did your disease 
\\ &interfere with your ability to do chores around the home?
 \\
$DA\_work$ & In the past seven days, how much did your disease 
\\ &interfere with your ability to perform well at work or school?
\\
$DA\_travel$ & In the past seven days, how much did your disease
\\ &interfere with your ability to travel more than one hour away from home?
 \\
$DA\_sleep$ & In the past seven days, how much did your disease 
\\ &interfere with your sleep?
 \\
$DA\_planning$ & In the past seven days, how much did your disease
\\ & make it difficult to plan several days ahead?
 \\
$DA\_leavinghome$ & In the past seven days, how much did your disease
\\ &make leaving home difficult?
 \\
$DA\_sex$ & In the past seven days, how much did your disease\\ & make you less interested in sex?
\\
$feeling\_alone$ & In the past seven days, how often did you feel alone?
 \\
$feeling\_angry$ & In the past seven days, how often did you feel angry?
\\
$feeling\_depressed$ & In the past seven days, how often did you feel depressed?
 \\
$feeling\_embarrased$ & In the past seven days, how often did you feel embarrassed?
 \\
$feeling\_frustrated$ & In the past seven days, how often did you feel frustrated?
 \\
$feeling\_scared$ & In the past seven days, how often did you feel scared? \\
$feeling\_worried$ & In the past seven days, how often did you feel worried? \\
$feeling\_nocontrol$ & In the past seven days, how often did you feel you had no control of your life?
\\
$bm\_leak$ & In the past 24 hours did stool, blood or liquid leak
\\ &out before you reached a toilet?  If so, how often?
 \\
$bm\_mucus$ & In the past 24 hours, did you have mucus (white material)
\\ &in your bowel movements?  If so, how often?
 \\
\bottomrule
\end{tabular}
\end{scriptsize}

\end{table}




\end{appendices}



\end{document}